\def\paperlanguage{}
\pdfoutput=1 

\documentclass[letterpaper, 10 pt, conference]{ieeeconf}  

\usepackage{bm}
\usepackage{cite}
\usepackage{flushend}
\include{preamble}

\IEEEoverridecommandlockouts                              

\title{\LARGE \textbf
  {
    \switchlanguage%
    {%
      Task-specific Self-body Controller Acquisition by Musculoskeletal Humanoids: Application to Pedal Control in Autonomous Driving
    }%
    {%
      筋骨格ヒューマノイドにおけるタスク特化した自己身体制御の獲得: 自動運転におけるペダル操作への応用
    }%
  }
}

\author{Kento Kawaharazuka$^{1}$, Kei Tsuzuki$^{1}$, Shogo Makino$^{1}$, Moritaka Onitsuka$^{1}$\\Koki Shinjo$^{1}$, Yuki Asano$^{1}$, Kei Okada$^{1}$, Koji Kawasaki$^{2}$, and Masayuki Inaba$^{1}$
  \thanks{$^{1}$ The authors are with the Department of Mechano-Informatics, Graduate School of Information Science and Technology, The University of Tokyo, 7-3-1 Hongo, Bunkyo-ku, Tokyo, 113-8656, Japan.
    {\texttt\small [kawaharazuka, tsuzuki, makino, onitsuka, shinjo, asano, k-okada, inaba]@jsk.t.u-tokyo.ac.jp}
  }
  \thanks{$^{2}$ The author is associated with TOYOTA MOTOR CORPORATION.
    {\texttt\small koji\_kawasaki@mail.toyota.co.jp}
  }
}
\begin{document}

\maketitle
\thispagestyle{empty}
\pagestyle{empty}

\begin{abstract}
  \switchlanguage%
  {%
    The musculoskeletal humanoid has many benefits that human beings have, but the modeling of its complex flexible body is difficult.
    Although we have developed an online acquisition method of the nonlinear relationship between joints and muscles, we could not completely match the actual robot and its self-body image.
    When realizing a certain task, the direct relationship between the control input and task state needs to be learned.
    So, we construct a neural network representing the time-series relationship between the control input and task state, and realize the intended task state by applying the network to a real-time control.
    In this research, we conduct accelerator pedal control experiments as one application, and verify the effectiveness of this study.
  }%
  {%
    筋骨格ヒューマノイドは様々な生物規範型の利点を有すると同時に, その複雑で柔軟な身体構造はモデリングが困難である.
    これまで関節と筋の非線形な関係を実機センサを用いて獲得していく手法を開発してきたが, 自己身体像を実機に完全に一致させることは困難であった.
    よって, あるタスクを実現する際は, そのタスクに特化した形で制御入力とタスク状態の関係を学習する必要がある.
    そこで, 制御入力とタスク状態の時系列関係を記述するネットワークを構成し, これを学習させ制御に応用することで, 所望のタスク状態を実現する.
    本研究では, 筋骨格ヒューマノイドによる車のペダル操作を例に実機実験を行い, 本手法の有効性を確認する.
  }%
\end{abstract}

\section{INTRODUCTION}\label{sec:introduction}
\begin{figure}[t]
  \centering
  \includegraphics[width=0.9\columnwidth]{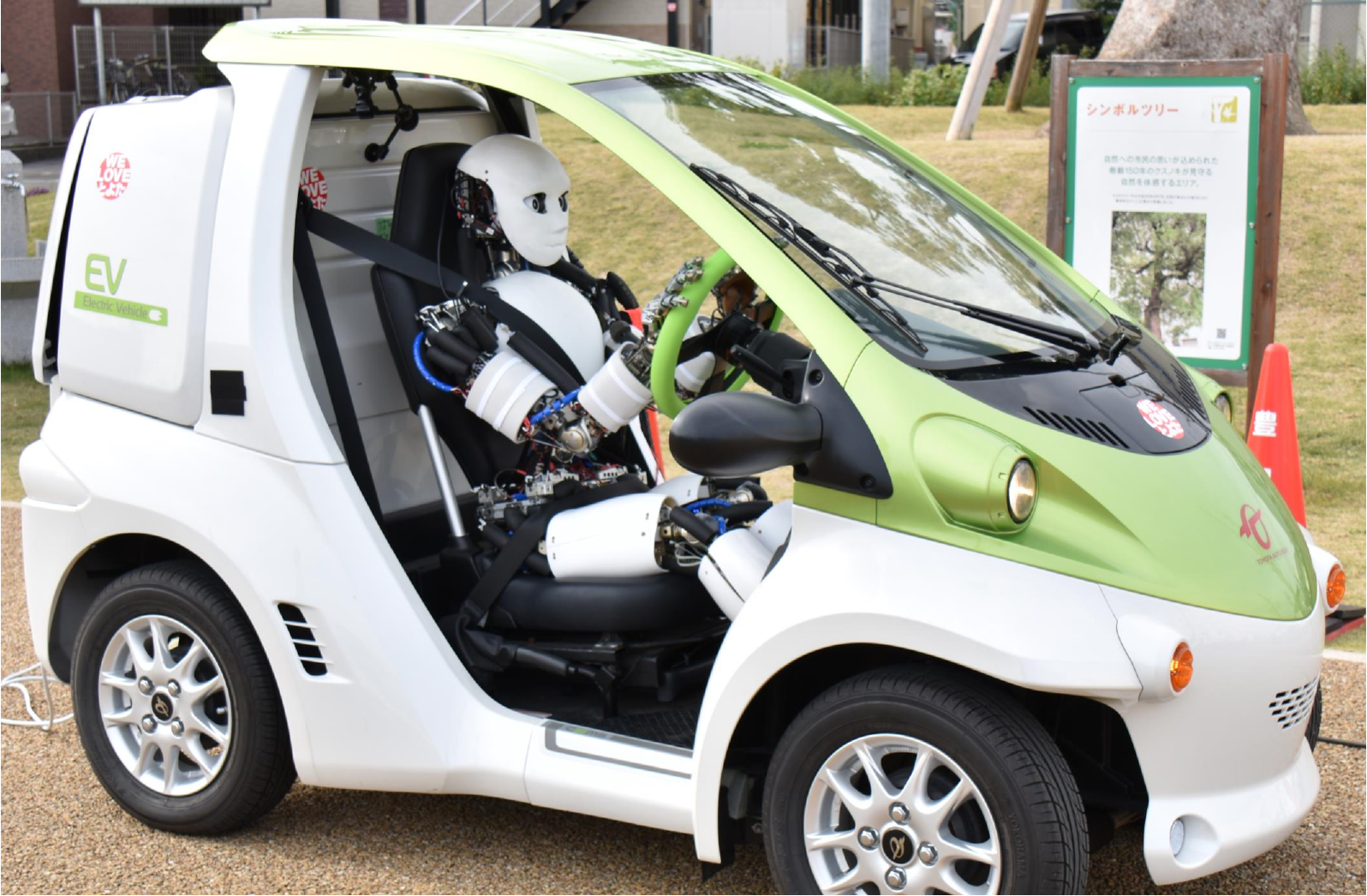}
  \caption{Autonomous driving by the musculoskeletal humanoid, Musashi.}
  \label{figure:musashi-driving}
  \vspace{-3.0ex}
\end{figure}

\switchlanguage%
{%
  The musculoskeletal humanoid \cite{wittmeier2013toward, nakanishi2013design, asano2016kengoro, jantsch2013anthrob} has many benefits such as the flexible under-actuated spine, ball joints without singular points, fingers with multiple degrees of freedom (multi-DOFs), variable stiffness control using redundant muscles, etc.
  At the same time, the modeling of its complex flexible body structure is difficult.
  To solve the problem, human-like reflex controls \cite{asano2013loadsharing, kawaharazuka2017antagonist} and body image acquisition methods \cite{kawaharazuka2018online, kawaharazuka2018bodyimage} have been developed.
  While the former is a temporary expedient to absorb the modeling error, the latter updates the self-body image online using the sensor information of the actual robot.
  However, even if we use the self-body image acquisition, we cannot completely match the actual robot and its self-body image.
  Thus, if we implement a control realizing a certain task in accordance with its modeling, the robot is hardly able to realize the task accurately.

  In this study, we develop a method to realize a certain task more accurately, by acquiring the self-body controller specializing in the specific task.
  By using this method, we do not need to tune the controller manually, and the robot can acquire the optimal self-body controller from a few motion data.

  The contributions of this study are as below.
  \begin{itemize}
    \item Discussion of the relationship between control input and state in musculoskeletal humanoids.
    \item The structure of DDC-Net (Dynamic Direct Control Network) representing the time-series relationship between control input and task state.
    \item The real-time control method realizing the intended task state using DDC-Net
    \item The evaluation of this method by accelerator pedal operation experiments, for autonomous driving by the musculoskeletal humanoid (\figref{figure:musashi-driving}).
  \end{itemize}

}%
{%
  筋骨格ヒューマノイド\cite{wittmeier2013toward, nakanishi2013design, asano2016kengoro}は, 柔軟で劣駆動な背骨や特異点のない球関節, 多自由度な手と指の構造や冗長な筋肉による可変剛性制御等の利点を有する.
  同時に, その複雑で柔軟な身体構造はモデリングが困難であり, これまで\cite{asano2013loadsharing}や\cite{kawaharazuka2017antagonist}等の反射に基づく制御手法, \cite{kawaharazuka2018online}や\cite{kawaharazuka2018bodyimage}等の自己身体像獲得手法が開発されてきた.
  前者は筋の反射によって対処療法的にモデリングの誤差を吸収するのに対して, 後者は自己身体のモデル自体を実機センサデータからのオンライン学習によって更新していく.
  しかし, 後者の方法を用いても, 実機と自己身体像の誤差を完全に無くすことはできない.
  そのため, ある特定のタスクを実現する制御をモデル通りに実装しても, 正しくそのタスクを実現できるとは限らない.

  そこで, 本研究では特定のタスクに特化して自己身体の制御法を学習させることで, そのタスクをより正確に実行できる手法を開発する.
  つまり, あるタスクを実現するためのコントローラをチューニングする必要がなくなり, 少ない実機の動作データから最適なコントローラを獲得することができる.

  本研究の詳細なコントリビューションを以下にまとめる.
  \begin{itemize}
      \item 筋骨格ヒューマノイドにおける制御入力と状態の関係と, これまでの自己身体像獲得に関する考察
      \item 時系列関係を考慮した制御入力とタスク状態の関係を記述するDDC-Net (Dynamic Direct Control Network)の構成
      \item DDC-Netを用いた, 動的に所望のタスクを実現可能な制御手法
      \item 筋骨格ヒューマノイドにおける自動運転のためのペダル操作実機実験による評価
  \end{itemize}

  以下ではまず, 筋骨格ヒューマノイドの身体構造の要約, 本研究と自己身体像獲得との関係, ヒューマノイドにおけるペダル操作について述べる.
  次に, 本研究の提案手法であるDDC-Netの構成と, DDC-Netを用いた所望の制御状態実現手法について述べる.
  最後に, 筋骨格ヒューマノイドによる車のペダル操作を例に取り\figref{figure:musashi-driving}, 実機実験により本手法の有効性を確認する.
}%

\section{Musculoskeletal Humanoids, Self-body Image Acquisition, and Autonomous Driving by Humanoids} \label{sec:musculoskeletal-upper-limb}
\subsection{Musculoskeletal Humanoids} \label{subsec:musculoskeletal-humanoids}
\begin{figure}[t]
  \centering
  \includegraphics[width=1.0\columnwidth]{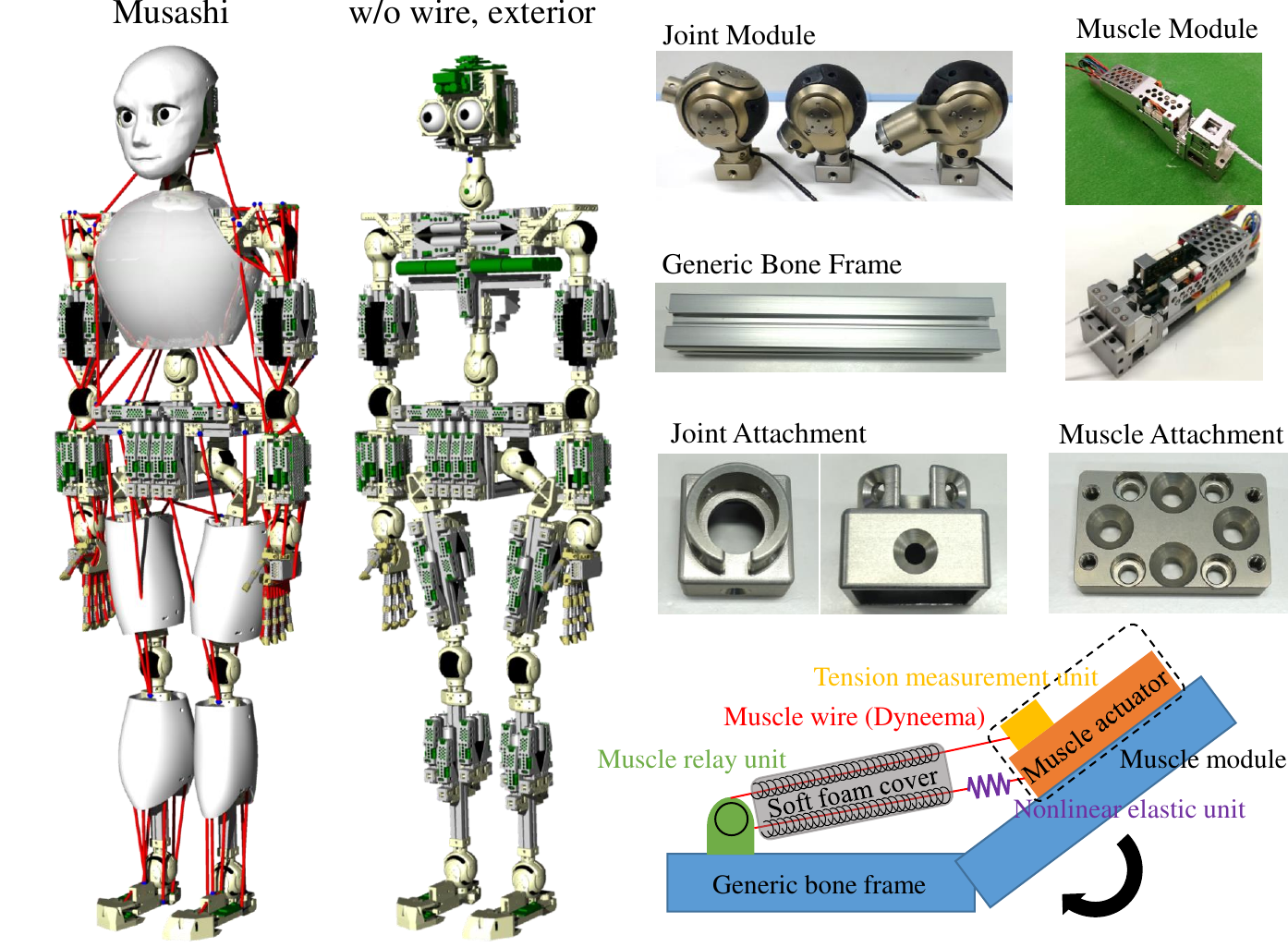}
  \vspace{-3.0ex}
  \caption{Details of the musculoskeletal humanoid, Musashi.}
  \label{figure:musculoskeletal-humanoid}
  \vspace{-3.0ex}
\end{figure}

\switchlanguage%
{%
  We show details of the musculoskeletal humanoid Musashi \cite{kawaharazuka2019musashi} used in this study (\figref{figure:musculoskeletal-humanoid}).
  This is a successor of Kengoro \cite{asano2016kengoro}, and is a modularized musculoskeletal humanoid platform composed of only joint modules, muscle modules, generic bone frames, and a few attachments.
  The muscles of its upper limb are composed of muscle wires using Dyneema and nonlinear elastic elements using O-ring, and there are no nonlinear elastic elements in its lower limb.
  Dyneema elongates like a spring, and the modeling of its friction, hysteresis, etc., is very difficult.
  In this study, we use only the ankle joint of Musashi.
}%
{%
  \figref{figure:musculoskeletal-humanoid}に本研究で用いる筋骨格ヒューマノイドMusashi \cite{kawaharazuka2019musashi}の詳細を示す.
  \cite{asano2016kengoro}の後継機であり, 全身が筋モジュール・関節モジュール・汎用骨格・少数のアタッチメントのみによって構成されたモジュラー型筋骨格ヒューマノイドである.
  上半身の筋肉はダイニーマによる筋ワイヤとOリングを用いた非線形弾性要素からなっており, 下半身はダイニーマのみで非線形弾性要素はない.
  ダイニーマはバネのように伸び, 500[N]で約8-10[mm]程度の伸びが発生するが, ヒステリシス等, そのモデリングは難しい.
  本研究では基本的にMusashiの足首関節のみを用いる.
}%

\subsection{Self-body Image Acquisition} \label{subsec:bodyimage-acquisition}
\begin{figure}[t]
  \centering
  \includegraphics[width=0.85\columnwidth]{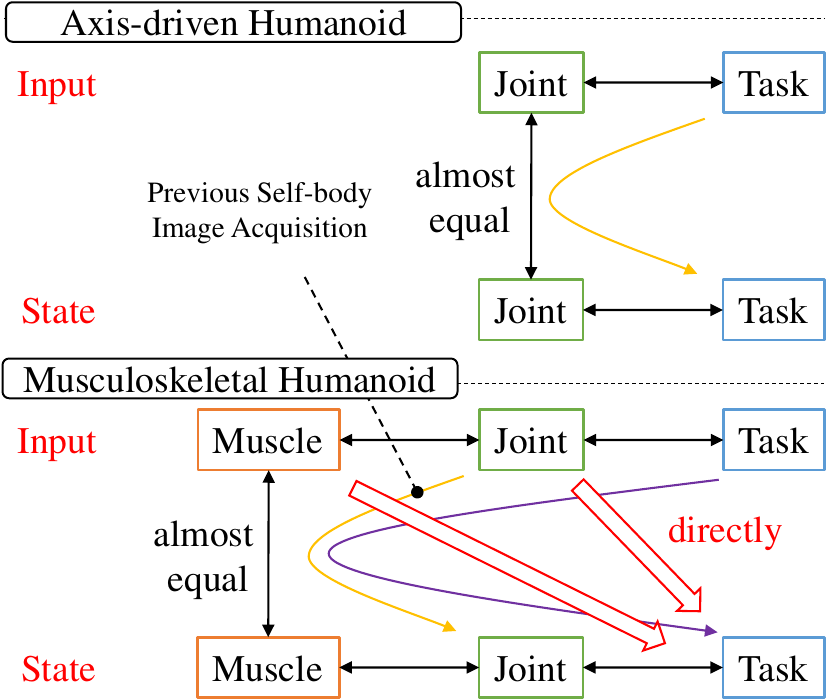}
  \caption{Comparison of robot controls between musculoskeletal humanoids and ordinary axis-driven humanoids.}
  \label{figure:control-comparison}
  \vspace{-1.0ex}
\end{figure}

\switchlanguage%
{%
  We compare body controls between the ordinary axis-driven humanoid \cite{hirai1998asimo} and the musculoskeletal humanoid in \figref{figure:control-comparison}.
  In this comparison, we define control input and state for muscle space, joint space, and task space (exempting operational space), respectively.

  Regarding the axis-driven humanoid, encoders are equipped with joints, and we execute a feedback control to match the target joint state and encoder value.
  So, in this system, we can almost completely match the control input and state.
  When using task space as control input, the control input in joint space is decided from the task input, the joint state follows it, and the task state changes.
  In this situation, the control input and state in joint space are almost equal, but the control input and state in task space will have some errors, because the task space cannot be handled directly.
  The task failures in DARPA Robotics Challenge \cite{darpa2015drc} are similar to this phenomenon.

  On the contrary, the musculoskeletal humanoid has a more complex control structure with an additional layer of muscle space.
  In this system, the joint space of the axis-driven humanoid and the muscle space of the musculoskeletal humanoid work almost equally, and the control input and state in those spaces match almost completely.
  Thus, to accurately realize the control input in joint space in the musculoskeletal humanoid is as difficult as to realize the control input in task space in the axis-driven humanoid.
  There are many previous studies that attempts to solve this problem.
  In \cite{ookubo2015learning}, the authors construct joint-muscle mapping offline using a table of joint angles and muscle lengths of the actual robot.
  In \cite{kawaharazuka2018online}, the authors express joint-muscle mapping by a neural network, and update it online using vision.
  In \cite{kawaharazuka2018bodyimage}, the authors update the relationship between joint angles, muscle tensions, and muscle lengths online, considering the body tissue softness unique to the musculoskeletal humanoid.
  By using these methods, the musculoskeletal humanoid can realize the intended joint angles to a certain degree, and realize object manipulation.
  However, these methods end in the learning of the relationship between joint space and muscle space, and to handle the control input in task space is more difficult.
  For example, these method cannot control a trajectory of a ball when thrown, and control the car velocity when driving.

  Thus, in this study, we develop a novel control method by directly representing the relationship between control input and task state.
  By using this method, the error intermediating multiple layers are absorbed, and the robot becomes to able to realize the task as intended.
  Also, while the previous studies \cite{ookubo2015learning, kawaharazuka2018online, kawaharazuka2018bodyimage} only handle the static state, we control the robot dynamically considering the time-series information.
  There are previous studies such as EMD Net \cite{tanaka2018emd} and SE3-Pose-Nets \cite{byravan2018se3}, but these studies do not handle the time-series information.
  Also in \cite{kawaharazuka2019dynamic}, there is no time-series output of task state and it does not consider regulations of control input.
}%
{%
  \figref{figure:control-comparison}に, 筋骨格ヒューマノイドと, 通常の軸駆動型ヒューマノイド\cite{hirai1998asimo}における身体制御の比較を示す.
  ここでは, 筋空間・関節空間・タスク空間(作業空間は取り除いている)の分類と, 制御入力・状態の分類を設けている.

  軸駆動型ヒューマノイドにおいては, 関節にエンコーダが備えられており, 関節状態を見ながら, 関節の制御入力に値を合わせるようなフィードバックをかける.
  そのため, 関節空間では制御入力と状態がほとんど同じになる.
  タスク空間を入力とした場合, そのタスクを実行する関節制御入力が決まり, 関節状態がそれに追従し, 行いたいタスクの状態が変化していく.
  関節空間においては制御入力と状態がほとんど同じであるが, 一段上層のタスク空間では制御入力と状態に間に誤差が溜まっていく.
  DARPA Robotics Challenge \cite{darpa2015drc}でタスクを失敗し転倒する事態も同じ現象と言える.

  これに対して, 人間のような筋骨格構造を持つ筋骨格ヒューマノイドは, 筋空間という層が一段増えた, より複雑な制御構造を有している.
  ここでは, 軸駆動型ヒューマノイドにおける関節空間と筋骨格ヒューマノイドにおける筋空間が最下層で同じ働きであり, 制御入力と状態がほぼ一致する.
  よって, 筋骨格ヒューマノイドにおいて関節角度を制御入力として, その関節角度状態を正確に実現するのは, 軸駆動型ヒューマノイドにおけるタスク空間と同じく, 難しい問題となる.
  この問題に対して, これまで多くの試みが成されている.
  \cite{ookubo2015learning}では, 実機の関節角度と筋長のテーブルを用いて, オフラインで関節-筋空間マッピングを生成している.
  \cite{kawaharazuka2018online}では, \cite{ookubo2015learning}をオンラインで実行するため, 関節-筋空間マッピングをニューラルネットワークを用いて表現し, 視覚を用いて更新する手法が開発されている.
  \cite{kawaharazuka2018bodyimage}では, \cite{kawaharazuka2018online}を発展させ, 筋骨格ヒューマノイド特有の身体組織の柔軟性を考慮した, 関節角度・筋張力・筋長の関係をオンラインに更新する手法が開発されている.
  これらの手法によって, 意図した関節角度をある程度実現し, 物体を掴むような動作が可能となってきた.
  しかしこれらは, 関節と筋の関係の学習に終始し, さらに一段層を深くし, タスク空間を制御入力とした場合, 層ごとに誤差が溜まっていくため, 問題はさらに難しくなる.
  例えば, ボールを投げるためにボールの速度や軌道を制御したり, ペダル操作で車速を制御したり, といった動作である.

  そこで本研究では, 筋骨格ヒューマノイドにおいて, タスク空間から関節空間, 関節空間から筋空間という層を介した方法ではなく, 制御入力から直接タスク状態を記述し, 制御する手法を開発する.
  これにより, 層を介すことによる誤差を吸収し, より意図した通りのタスク実行を可能とすることができる.
  また, \cite{ookubo2015learning, kawaharazuka2018online, kawaharazuka2018bodyimage}では静的な状態のみを扱っていたが, 本研究では時系列情報を考慮し, 動的な制御を可能とする.
  その他先行研究として, EMD Net \cite{tanaka2018emd}やSE3-Pose-Nets \cite{byravan2018se3}等が存在するが, どれも時系列情報を扱った動的動作の実現はできていない.
}%

\begin{figure}[t]
  \centering
  \includegraphics[width=1.0\columnwidth]{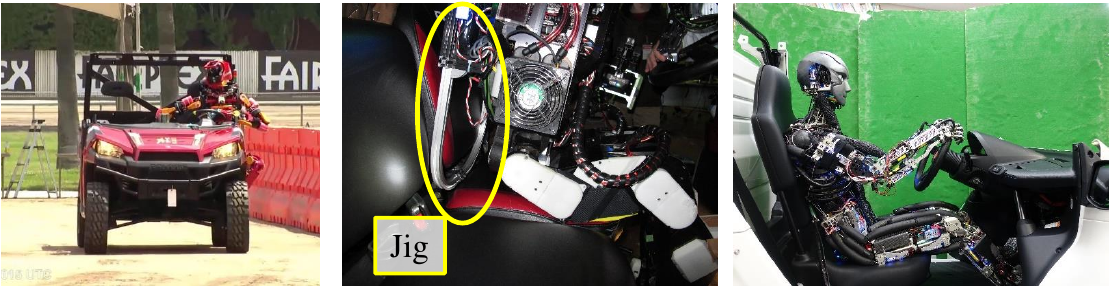}
  \vspace{-3.0ex}
  \caption{Comparison of autonomous driving between using musculoskeletal humanoids and ordinary axis-driven humanoids. In DARPA robotics challenge \cite{darpa2015drc}, jaxon \cite{kojima2015jaxon} needed a jig to sit on the seat.}
  \label{figure:humanoid-driving}
  \vspace{-3.0ex}
\end{figure}

\begin{figure*}[t]
  \centering
  \includegraphics[width=1.74\columnwidth]{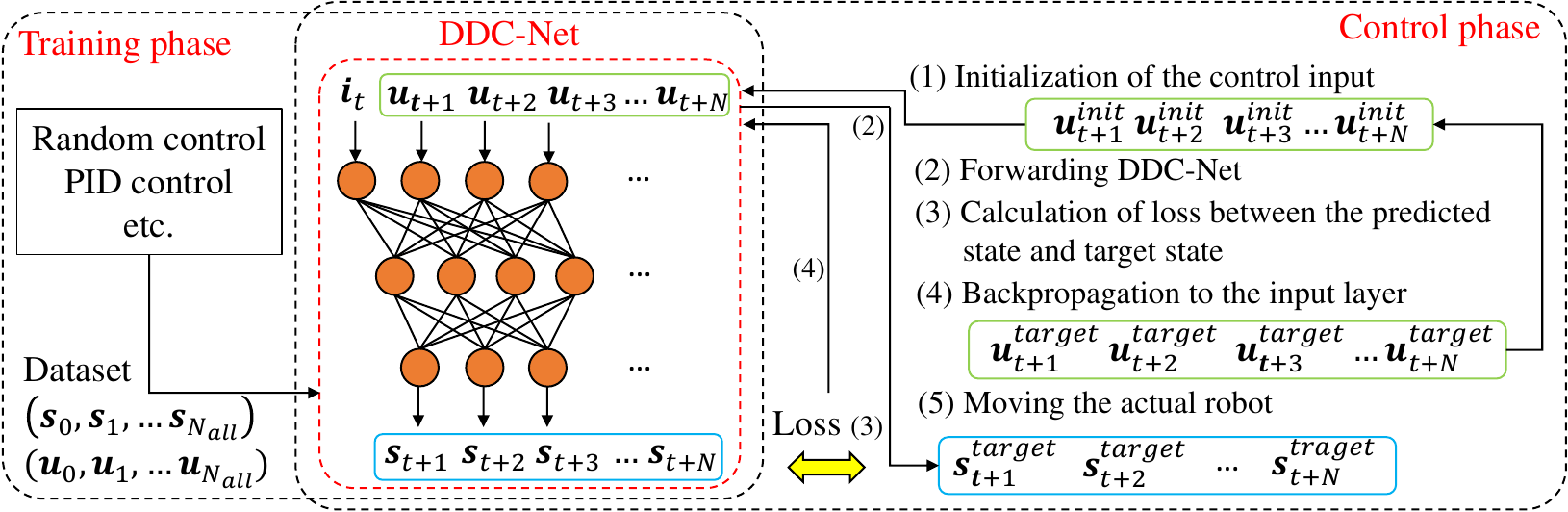}
  \caption{Overview of the task-specific self-body controller acquisition method.}
  \label{figure:ddcn-overview}
  \vspace{-3.0ex}
\end{figure*}

\subsection{Autonomous Driving by Musculoskeletal Humanoids} \label{subsec:humanoid-self-driving}
\switchlanguage%
{%
  In the driving task of DRC \cite{darpa2015drc}, the ordinary axis-driven humanoid JAXON \cite{kojima2015jaxon} needed a jig to sit on the car seat, as shown in \figref{figure:humanoid-driving}.
  On the contrary, the musculoskeletal humanoid has human-like proportions, so it can move in the car and adapt its flexible body to the environment.
  Therefore, studying autonomous driving by the musculoskeletal humanoid is important, and we handle the accelerator pedal operation in this study.
  In our experiments, we operate the accelerator pedal by the ankle pitch joint to control the car velocity.
  Also, the feedback frequency of human pedal operation is slow, the frequency being above about 3.5 Hz \cite{sande2012pedal}, and we control the robot by 5 Hz in this study.
}%
{%
  DARPA Robotics Challenge \cite{darpa2015drc}の運転タスクにおいて, 通常の軸駆動型ヒューマノイドであるJAXON\cite{kojima2015jaxon}は車のシートに座るためにジグを要した(\figref{figure:humanoid-driving}).
  それに対して, 筋骨格ヒューマノイドは人体のプロポーションを模倣しているがゆえに, 人間のために作られた車体の中に容易に入ることができ, その柔軟な身体を環境に馴染ませることができる.
  そのため, 筋骨格ヒューマノイドを運転操作に用いることは有用であると考え, 本研究では運転において重要なペダル操作を扱う.
  ここでは基本的に, 足首のpitch軸によってアクセルを操作し, 車体の速度を制御する.
  また, 人間のペダル操作における感覚フィードバックは遅く, 3.5 Hz以上と言われており\cite{sande2012pedal}, 本研究では5 Hzの周期で制御を行う.
}%

\section{Task-specific Self-body Controller Acquisition} \label{sec:selfbody-controller}
\switchlanguage%
{%
  We show the whole system of the task-specific self-body controller acquisition method in \figref{figure:ddcn-overview}.
  We will explain the network structure of DDC-Net, its training phase, and the control phase using it.
}%
{%
  タスク特化型自己身体制御獲得手法の全体の構成を\figref{figure:ddcn-overview}に示す.
  DDC-Netのネットワーク構造, 訓練フェーズ, 制御フェーズについてそれぞれ詳細を述べる.
}%

\subsection{Network Structure} \label{subsec:network-structure}
\switchlanguage%
{%
  DDC-Net in this study is equal to the function $\bm{f}$ as below:
  \begin{align}
    \begin{bmatrix}
      \bm{s}_{t+1}^T & \bm{s}_{t+2}^T & \cdots & \bm{s}_{t+N}^T
    \end{bmatrix}
    ^T=\;\;\;\;\;\;\;\;\;\;\;\;\;\;\;\;\;\;\;\;\;\;\;\;&\notag\\\bm{f}(
    \begin{bmatrix}
      \bm{i}_{t}^T & \bm{u}_{t+1}^T & \bm{u}_{t+2}^T & \cdots & \bm{u}_{t+N}^T
    \end{bmatrix}
    ^T)& \label{equation:ddc-net}
  \end{align}
  where $\bm{s}$ is the task state with $N_s$ dimensions, $\bm{i}$ is the initial state with $N_i$ dimensions, $N$ is the length of time-series information to consider, and $\bm{u}$ is the control input with $N_u$ dimensions.
  By feeding the time-series control input to the initial state, we can predict all the time-series task states until $N$ steps ahead.
  We generally set $\bm{i}_{t}=\bm{s}_{t}$, but we can increase the initial state information.

  Although any network structures that express \equref{equation:ddc-net} are fine, we use the network with 5 fully-connected layers in this study.
  For all the layers except the last layer, we apply batch normalization \cite{ioffe2015batchnorm} after each layer, and use sigmoid as activation functions.
  The dimension of the input layer is $N_i+N{\times}N_u$, and the dimension of the output layer is $N{\times}N_s$.

  In our pedal operation experiment, we set $N=30$, $N_s=1$ (car velocity), $N_i=4$ (car velocity, car acceleration, joint angle of ankle pitch, and joint velocity of ankle pitch), $N_u=1$ (joint angle of ankle pitch), and the number of units of the hidden layers are $\{80, 50, 20\}$.
}%
{%
  本研究におけるDDC-Netは, 以下の関数$\bm{f}$に等しい.
  \begin{align}
    \begin{bmatrix}
      \bm{s}_{t+1}^T & \bm{s}_{t+2}^T & \cdots & \bm{s}_{t+N}^T
    \end{bmatrix}
    ^T=\;\;\;\;\;\;\;\;\;\;\;\;\;\;\;\;\;\;\;\;\;\;\;\;&\notag\\\bm{f}(
    \begin{bmatrix}
      \bm{i}_{t}^T & \bm{u}_{t+1}^T & \bm{u}_{t+2}^T & \cdots & \bm{u}_{t+N}^T
    \end{bmatrix}
    ^T)& \label{equation:ddc-net}
  \end{align}
  ここで, $\bm{s}$は$N_s$の次元を持つタスクの状態, $\bm{i}$は$N_i$の次元を持つ初期状態, $N$は考慮する時系列情報の長さを表す定数, $\bm{u}$は$N_u$の次元を持つ制御入力を表す.
  初期状態に, $N$個の制御入力の時系列情報を与えることによって, その先の$N$個のタスク状態の時系列情報を予測する.
  $\bm{i}_{t}=\bm{s}_{t}$が基本であるが, 初期状態の情報を増やす場合も考えられる.

  このネットワーク構造は何を用いても良いが, 本研究では, 5層の全結合層からなる構造を用いてる.
  最終層以外の前層対しては, 全結合層の後にBatch Normalization\cite{ioffe2015batchnorm}を適用し, 活性化関数としてはsigmoidを用いている.
  入力層の次元数は, $N_i+N{\times}N_u$であり, 出力層の次元数は$N{\times}N_s$となる.

  本研究のペダル操作では, $N=30$, $N_s=1$(車の速度), $N_i=4$(車の速度, 車の加速度, 足首ピッチ軸の関節角度, 足首ピッチ軸の関節角速度), $N_u=1$(足首ピッチ軸の関節角度)となっており, 隠れ層のユニット数は, $\{80, 50, 20\}$としている.
}%

\subsection{Training Phase} \label{subsec:training-phase}
\switchlanguage%
{%
  In the training phase, we first accumulate the motion data when controlling the robot by random control input.
  We can obtain the motion data of $\bm{s}$ and $\bm{u}$, convert them to the forms of $(\bm{i}_{t}, \bm{u}_{t+1}, \bm{u}_{t+2}, \cdots, \bm{u}_{t+N})$ and $(\bm{s}_{t+1}, \bm{s}_{t+2}, \cdots, \bm{s}_{t+N})$, and train DDC-Net (we make $\bm{i}$ from $\bm{s}$, $\bm{u}$, their differential values, etc.).
  In this study, we use Mean Squared Error (MSE) as the loss function, and Adam \cite{kingma2015adam} as the update method.
  We use the 1/5 of the data as test data, and use the model with the lowest loss in 100 epoch for the control phase.
}%
{%
  訓練フェーズにおいては, まず, ロボットをランダムな制御入力等によって動かした際のデータを蓄積する.
  動作データとして$\bm{s}$と$\bm{u}$が得られるため, これを$(\bm{i}_{t}, \bm{u}_{t+1}, \bm{u}_{t+2}, \cdots, \bm{u}_{t+N})$と$(\bm{s}_{t+1}, \bm{s}_{t+2}, \cdots, \bm{s}_{t+N})$のようにネットワークに適する形に変換し, DDC-Netを学習させる($\bm{i}$は$\bm{s}$や$\bm{u}$, それらの微分値等から作成する).
  本研究では, 損失関数として平均二乗誤差(MSE, Mean Squared Error)を用い, 更新則にはAdam \cite{kingma2015adam}を用いる.
  全データの1/5をtestに用い, 100 epoch学習させた際に最もtestのlossが小さなモデルを制御フェーズでは使用する.
}%

\subsection{Control Phase} \label{subsec:control-phase}
\switchlanguage%
{%
  There are five steps in the control phase as below.
  \begin{enumerate}
      \item Get the current task state and decide the initial value of the target control input.
      \item Feed them into DDC-Net.
      \item Calculate the loss between the predicted task state from DDC-Net and target task state.
      \item Backpropagate the loss to the initial value of the target control input.
      \item Repeat (2)--(4), and finally send the calculated target control input to the robot.
  \end{enumerate}

  In (1), we decide the current state $\bm{i}_{t}$ and initial value of the target control input $\bm{u}^{init}_{seq}=\{\bm{u}^{init}_{t+1}, \bm{u}^{init}_{t+2}, \cdots, \bm{u}^{init}_{t+N}\}$.
  We set $\bm{u}^{init}_{seq}$ as $\{\bm{u}^{pre}_{t+1}, \bm{u}^{pre}_{t+2}, \cdots, \bm{u}^{pre}_{t+N-1}, \bm{u}^{pre}_{t+N-1}\}$ which shifts the optimized target control input of the previous time  step  $\{\bm{u}^{pre}_{t}, \bm{u}^{pre}_{t+1}, \cdots, \bm{u}^{pre}_{t+N-1}\}$ and replicates the last term $\bm{u}^{pre}_{t+N-1}$.

  In (2)--(3), we first feed the obtained values from (1) into DDC-Net.
  Next, we calculate the loss ($L$) between the predicted task state $\bm{s}^{predicted}_{seq}$ and target task state $\bm{s}^{target}_{seq}$.
  In this study, we calculate the loss as below:
  \begin{align}
    L = \textrm{MSE}(\bm{s}^{predicted}_{seq}, \bm{s}^{target}_{seq}) + \alpha\textrm{AdjacentError}(\bm{u}^{init}_{seq}) \label{equation:loss}
  \end{align}
  where MSE is the Mean Squared Error, $\alpha$ is the weight of losses, AdjacentError is the MSE of each difference of target control inputs at adjacent time steps.
  We added the last term to decrease the changes in control input and obtain smooth control input.

  In (4), we update $\bm{u}^{init}_{seq}$ by backpropagating the loss, as below:
  \begin{align}
    \bm{g} &= dL/d\bm{u}^{init}_{seq} \\
    \bm{u}^{init}_{seq} &= \bm{u}^{init}_{seq} - \beta\frac{\bm{g}}{|\bm{g}|}
  \end{align}
  where $\bm{g}$ is the gradient of loss($L$) for $\bm{u}^{init}_{seq}$, and $\beta$ is the learning rate constant.
  By repeating (2)--(4), we update $\bm{u}^{init}_{seq}$.

  In (5), we send $\bm{u}^{target}_{t+1}$ of the finally calculated $\bm{u}^{target}_{seq}$ to the actual robot.

  We mentioned the overview as above, and we add some changes to calculate the target control input by dividing (1)--(4) to two stages.
  In the first stage, we add random noises from $-\delta{\bm{u}}_{batch}$ to $\delta{\bm{u}}_{batch}$ in $\bm{u}^{init}_{seq}$, construct $N_{batch}$ number of samples, and set the $N_{batch}+1$ number of samples, including the sample without noise, as the initial control input.
  We repeat (2)--(4) $N_{1}$ times, by setting $\beta=\beta_{1}$.
  In the second stage, we use the updated control input with the lowest loss in the first stage, as the initial control input.
  We set $\beta=\beta_{2}$ ($\beta_{2}<\beta_{1}$), repeat (2)--(4) $N_{2}$ times, and calculate the final $\bm{u}^{target}_{seq}$.
  Since $\bm{u}$ has the minimum and maximum values, we filter $\bm{u}$ to fit from $\bm{u}_{min}$ to $\bm{u}_{max}$.
  Also, before the first calculation, there are no values of $\bm{u}^{pre}_{seq}$, so we use $N_{batch}$ number of samples as $\bm{u}^{init}_{seq}$ by filling the same random value from $\bm{u}_{min}$ to $\bm{u}_{max}$ at every time.

  In our pedal operation experiment, we set $u_{min}=0$ [deg], $u_{max}=50$ [deg], $N_{batch}=10$, $N_{1}=10$, $N_{2}=20$, $\delta{u}_{batch}=5$ [deg], $\alpha=30.0$, $\beta_{1}=3.0$ [deg], and $\beta_{2}=0.5$ [deg].

  In this study, we use a joint angle as the control input.
  Therefore, after we convert the joint angle to the muscle length by \cite{kawaharazuka2018bodyimage}, we send it to the actual robot.
  There is no need of joint angle sensors.
}%
{%
  制御フェーズにおいては, 以下の5つの工程を実行する.
  \begin{enumerate}
      \item 現在のタスク状態の取得と, 指令制御入力の初期値決定
      \item DDC-Netの順伝播
      \item DDC-Netから出力された予測タスク状態と指令タスク状態のlossを計算
      \item 指令制御入力の初期値に対して誤差逆伝播
      \item (2)--(4)を繰り返し, 最終的な指令制御入力値をロボットに指令
  \end{enumerate}

  (1)においては, 現在の状態$\bm{i}_{t}$と指令制御入力の初期値$\bm{u}^{init}_{seq}=\{\bm{u}^{init}_{t+1}, \bm{u}^{init}_{t+2}, \cdots, \bm{u}^{init}_{t+N}\}$を決定する.
  ここで指令制御入力は, 前ステップで最適化され最終的にロボットに指令された$\{\bm{u}^{pre}_{t}, \bm{u}^{pre}_{t+1}, \cdots, \bm{u}^{pre}_{t+N-1}\}$を一つずらして最終ステップ複製した, $\{\bm{u}^{pre}_{t+1}, \bm{u}^{pre}_{t+2}, \cdots, \bm{u}^{pre}_{t+N-1}, \bm{u}^{pre}_{t+N-1}\}$を用いる.

  (2)--(3)について, まずはDDC-Netを順伝播する.
  その後, 出力された予測タスク状態$\bm{s}^{predicted}_{seq}$と指令タスク状態$\bm{s}^{target}_{seq}$のloss($L$)を計算する.
  本研究では以下のlossを用いる.
  \begin{align}
    L = \textrm{MSE}(\bm{s}^{predicted}_{seq}, \bm{s}^{target}_{seq}) + \alpha\textrm{AdjacentError}(\bm{u}^{init}_{seq}) \label{equation:loss}
  \end{align}
  ここで, MSEはMean Squared Error, $\alpha$は重みづけの係数, AdjacentErrorは制御入力のステップ間の値のMSEを表す.
  最終項は, ステップ間の制御入力の差を小さくすることで, 滑らかな制御入力を導出するための工夫である.

  (4)について, 最後に, lossから誤差逆伝播により$\bm{u}^{init}_{seq}$を以下のように更新する.
  \begin{align}
    \bm{g} &= dL/d\bm{u}^{init}_{seq} \\
    \bm{u}^{init}_{seq} &= \bm{u}^{init}_{seq} - \beta\frac{\bm{g}}{|\bm{g}|}
  \end{align}
  ここで, $\bm{g}$はloss($L$)の$\bm{u}^{init}_{seq}$に対する勾配, $\beta$は学習率を表す定数である.
  (2)--(4)の工程を繰り返して$\bm{u}^{init}_{seq}$を更新していく.

  (5)においては, 最終的に求まった$\bm{u}^{target}_{seq}$の$\bm{u}^{target}_{t+1}$を実機に指令する.

  上記が概要であるが, 本研究ではさらに, (1)--(4)の工程を二段階に分けて更新する工夫を施している.
  第一段回では, $\bm{u}^{init}_{seq}$に$-\delta{\bm{u}}_{batch}$から$\delta{\bm{u}}_{batch}$のランダムなノイズを加えて, $N_{batch}$個のサンプルを作成し, ノイズを入れないものも含め$N_{batch}+1$個の初期値をバッチとして入力する.
  (2)--(4)の工程を, $\beta=\beta_{1}$に設定して$N_{1}$回繰り返す.
  第二段階では, 第一段階で最終的に最もlossが小さかった指令制御入力を初期値として用いる.
  $\beta=\beta_{2}$ ($\beta_{2}<\beta_{1}$)に設定し, (2)--(4)の工程を$N_{2}$回繰り返し, 最終的な$\bm{u}^{target}_{seq}$を求める.
  $\bm{u}$には最小値最大値が存在するため, その都度$\bm{u}_{min}$から$\bm{u}_{max}$の間に値が収まるようにフィルタをかける.
  また, 最初の施行時には$\bm{u}^{init}_{seq}$が存在しないため, 全時間の$\bm{u}$を$\bm{u}_{min}$から$\bm{u}_{max}$のランダムな同一の値で埋めた$\bm{u}^{init}_{seq}$を$N_{batch}$個作成してバッチとして用いる.

  本研究のペダル操作においては, $u_{min}=0$ [deg], $u_{max}=50$ [deg], $N_{batch}=10$, $N_{1}=10$, $N_{2}=20$, $\delta{u}_{batch}=5$ [deg], $\alpha=30.0$, $\beta_{1}=3.0$ [deg], $\beta_{2}=0.5$ [deg]としている.

  本研究では関節角度を制御指令とし, その関節角度は\cite{kawaharazuka2018bodyimage}によって筋長に変換された後, 実際に実機に送られており, 関節角度センサを必要としない.
}%

\section{Pedal Control Experiments} \label{sec:experiment}
\switchlanguage%
{%
  First, we will explain the experimental setup.
  Next, we will conduct accelerator pedal control experiments using an ordinary PID control and extended PID control with the acceleration estimation.
  Although various model-based control methods have been developed, we cannot apply those due to the difficult modeling of this experiment, and used PID controllers.
  Finally, we will conduct the developed task-specific self-body controller acquisition method, and verify its effectiveness.
}%
{%
  まず, 本研究で用いる実験装置について説明する.
  次に, 通常のPID制御, 加速度推定を行ったPID制御によるペダル操作実験を行う.
  最後に, 本研究で提案したタスク特化型自己身体制御獲得手法を用いた実験を行う.
}%

\begin{figure}[t]
  \centering
  \includegraphics[width=1.0\columnwidth]{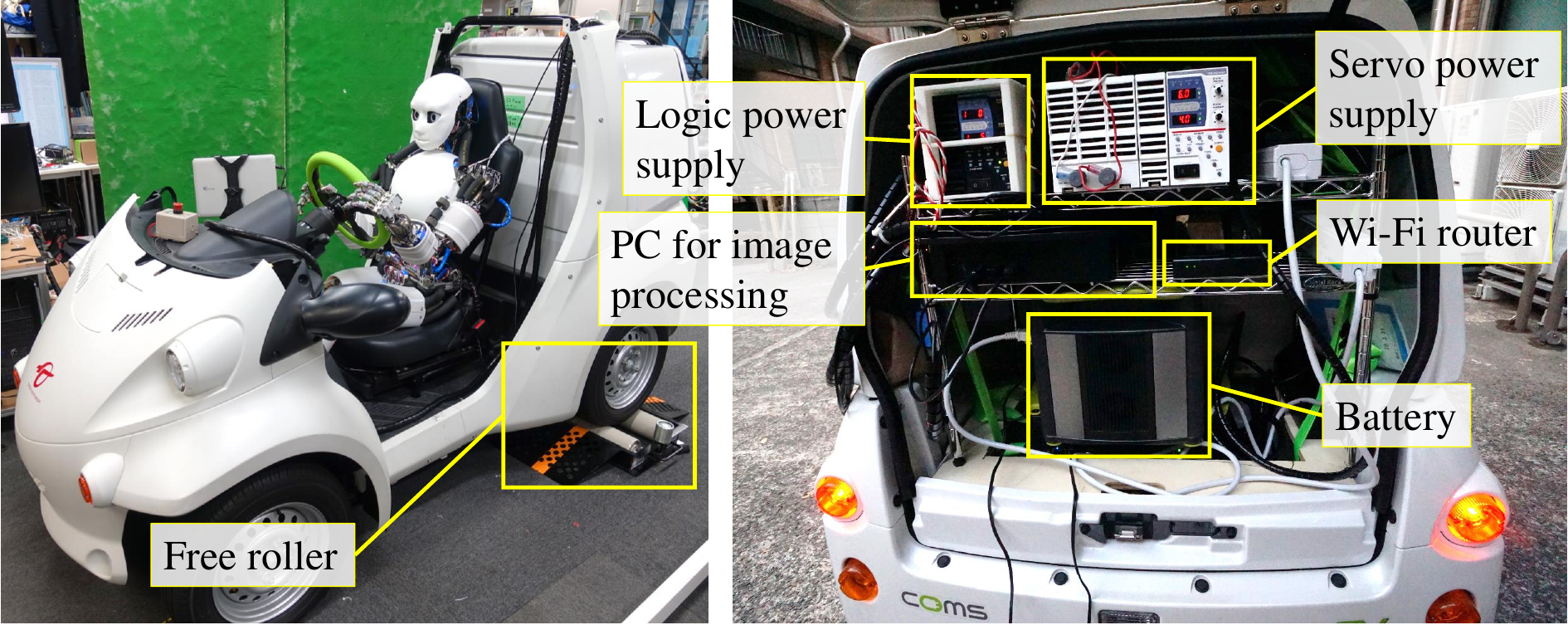}
  \vspace{-3.0ex}
  \caption{Experimental setup of COMS.}
  \label{figure:coms-setup}
  \vspace{-3.0ex}
\end{figure}

\subsection{Experimental Setup}
\switchlanguage%
{%
  The car used in this experiment is B.COM Delivery of extremely small EV COMS (Chotto Odekake Machimade Suisui) series.
  For the safety, the motor torque is limited to 5 Nm on its software, and the emergency stop button is equipped.

  The outside appearance is shown in the left figure of \figref{figure:coms-setup}.
  Although we should obtain the current car velocity by visual odometry, etc. from image information, we conduct experiments in the indoor environment for the safety, and we obtain the car velocity from CAN-USB from ROS.
  The rear drive wheels are on the free roller, and the front wheels are fixed by stoppers.

  Also, the inside appearance is shown in left figure of \figref{figure:coms-setup}.
  In the car, power supplies for the logic and motors of the robot, batteries, Wi-Fi rooter are equipped.
  We use special batteries for the power supply, but in the future, we examine to directly obtain the power from the electric vehicle COMS.
}%
{%
  本研究のペダル操作実験で用いる自動車は, トヨタ車体製の超小型EV コムス (COMS: Chotto Odekake Machimade Suisui)シリーズのB・COMデリバリーである.
  安全のため, モータトルクはソフトウェア上で5 Nmに制限されており, 非常停止ボタンが備えられている.

  実験装置の外観は\figref{figure:coms-outer}のようになっている.
  通常の運転であれば画像情報からのVisual Odometry等で車体の現在速度を取得するべきであるが, 安全のため室内環境で実験を行い, 車体速度はCAN-USBによりROSを介してCOMSのソフトウェアから得ている.
  駆動輪である後輪はフリーローラの上に載せており, 前輪は安全を期してストッパーにより固定されている.

  また, COMSの内部は\figref{figure:coms-inner}のようになっている.
  車体にはロジック・パワーの電源装置, バッテリー, WiFiルータが装備されている.
  本研究では別途バッテリーや電源装置を用意しているが, 最終的には, 電気自動車であるCOMSから電源を取ることを検討している.
}%

\begin{figure}[t]
  \centering
  \includegraphics[width=0.95\columnwidth]{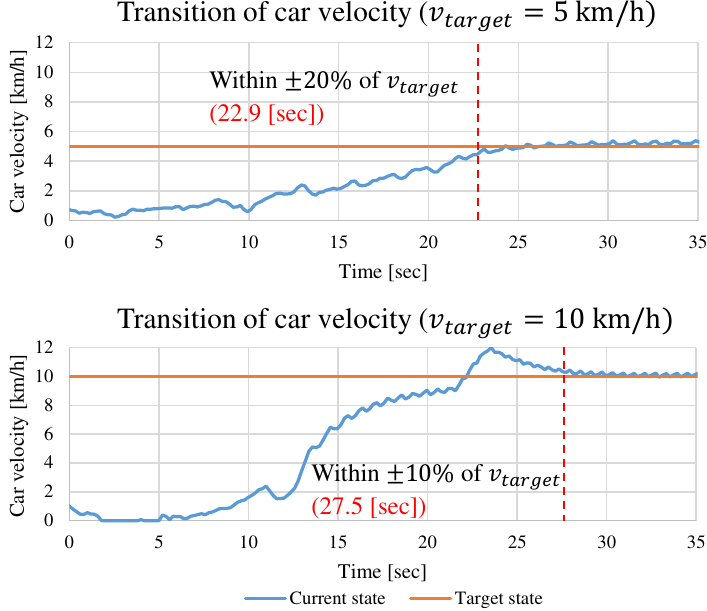}
  \caption{The result of the basic PID control (\textbf{PID1}).}
  \label{figure:pid-control1}
  \vspace{-3.0ex}
\end{figure}

\begin{figure}[t]
  \centering
  \includegraphics[width=0.95\columnwidth]{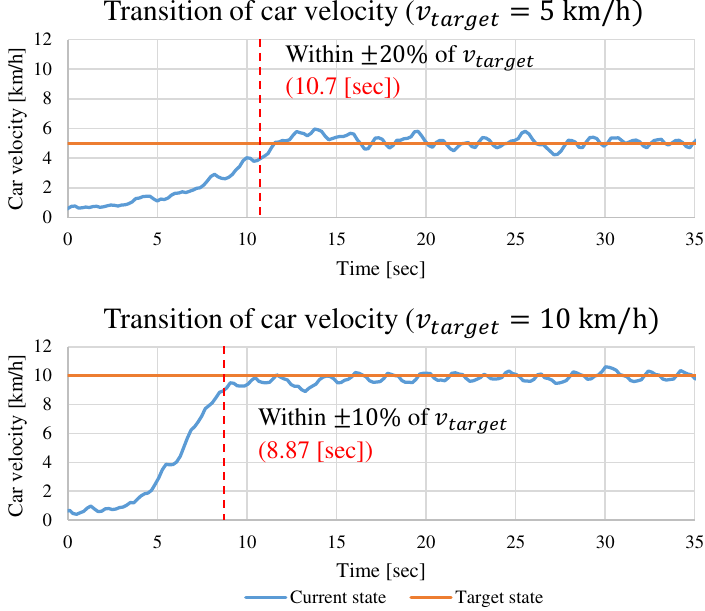}
  \caption{The result of the PID control with acceleration estimation (\textbf{PID2}).}
  \label{figure:pid-control2}
  \vspace{-3.0ex}
\end{figure}

\subsection{Basic Control}
\switchlanguage%
{%
  First, we show the result of the car velocity transition when using a basic PID control (we call it \textbf{PID1}) in \figref{figure:pid-control1}.
  $PID1$ is the control as shown below.
  \begin{align}
    e_{v}(t) &= v_{target}(t) - v(t)\\
    u &= \theta_{ankle} = k^{1}_{p}e_{v}(t) + k^{1}_{i}\int e_{v}(t) + k^{1}_{d}\dot{e_{v}}(t)
  \end{align}
  Where $k^{1}_{p}, k^{1}_{i}, k^{1}_{d}$ are P, I, D gain respectively, $v$ is the current car velocity, and $u=\theta_{ankle}$ is the joint angle of the ankle pitch as the control input.
  We conduct all experiments to track the constant target car velocity of 5 km/h or 10 km/h.
  We can see that \textbf{PID1} is slow to follow the target car velocity from \figref{figure:pid-control1}.
  If we define $T^{conv}_{A\%}$ as the first time that the error between the current and target car velocity falls within A\% from then, $T^{conv}_{20\%}=22.9$ [sec] when $v_{target}=5$ [km/h], and $T^{conv}_{10\%}=27.5$ [sec] when $v_{target}=10$ [km/h].
  Because the tracking control is difficult when the target car velocity is slow, we basically set $A=20$ when $v_{target}=5$, and $A=10$ when $v_{target}=10$ [km/h], for the experimental evaluation.

  We think that the reason of the slow tracking control by \textbf{PID1} is because the amount of pedal step actually effects not the car velocity directly but the car acceleration.
  So, we consider to estimate the target car acceleration and to execute PD control to the value (we call it \textbf{PID2}).
  In detail, we conduct a control as shown below.
  \begin{align}
    a_{target}(t) &= (v_{target}(t)-v(t))/t_{delay} \\
    a(t) &= (v(t)-v(t-1))/t_{interval} \\
    e_{a}(t) &= a_{target}(t)-a(t) \\
    u &= \int{(k^{2}_{p}e_{a}(t) + k^{2}_{d}\dot{e_{a}}(t))}
  \end{align}
  Where, $a_{target}$ is the target acceleration, $t_{delay}$ is the time delay until realizing the target acceleration, $t_{interval}$ is the control period, $a$ is the current car acceleration, and $k^{2}_{p}, k^{2}_{d}$ is the P and D gain, respectively.
  We show the result of \textbf{PID2} in \figref{figure:pid-control2}.
  We can see that $T^{conv}_{20\%}=10.7$ [sec] when $v_{target}=5$ [km/h], and $T^{conv}_{10\%}=8.87$ [sec] when $v_{target}=10$ [km/h].
  Compared to \textbf{PID1}, the time delay to follow the target velocity is improved in \textbf{PID2}.
}%
{%
  まず, 従来のような通常のPID制御(以降PID1と呼ぶ)を行った際の車体速度変化を\figref{figure:pid-control1}に示す.
  これは,
  \begin{align}
    e_{v}(t) &= v_{target}(t) - v(t)\\
    u &= \theta_{ankle} = k^{1}_{p}e_{v}(t) + k^{1}_{i}\int e_{v}(t) + k^{1}_{d}\dot{e_{v}}(t)
  \end{align}
  に基づいた制御であり, $k^{1}_{p}, k^{1}_{i}, k^{1}_{d}$はそれぞれPゲイン, Iゲイン, Dゲイン, $v$は現在車速, $u=\theta_{ankle}$は制御入力である足首pitchの関節角度を表す.
  $v_{target}$は指令速度であり, 全実験で5 km/h, 10 km/hの二種類の一定速度への追従実験を行っている.
  \figref{figure:pid-control1}からわかるように, PID1は指令速度への追従が遅い.
  それ以降指令速度からの誤差がA\%に収まる最初の時間を$T^{conv}_{A\%}$とすると, $v_{target}=5$ [km/h]のとき$T^{conv}_{20\%}=22.9$ [sec], $v_{target}=10$ [km/h]のとき$T^{conv}_{10\%}=27.5$ [sec]となっている.
  車速が遅いときは速度追従のが難しいため, $v_{target}=5$ [km/h]では$A=20$, $v_{target}=10$ [km/h]では$A=10$を評価の基本としている.

  PID1による速度追従が遅い理由は, アクセルペダルの踏み込み量は実際には車体の速度ではなく, 加速度項に大きく影響を与える量だからであると考える.
  そこで, 目標加速度を推定し, その加速度に対してPD制御を行うことを考える(以降PID2と呼ぶ).
  具体的には, 以下のような制御を行う.
  \begin{align}
    a_{target}(t) &= (v_{target}(t)-v(t))/t_{delay} \\
    a(t) &= (v(t)-v(t-1))/t_{interval} \\
    e_{a}(t) &= a_{target}(t)-a(t) \\
    u &= \int{(k^{2}_{p}e_{a}(t) + k^{2}_{d}\dot{e_{a}}(t))}
  \end{align}
  ここで, $a_{target}$は目標加速度, $t_{delay}$は目標速度を実現するまでの時間遅れ, $t_{interval}$は制御間隔, $a$は現在加速度, $k^{2}_{p}, k^{2}_{d}$はそれぞれPゲイン, Dゲインを表す.
  PID2による速度制御の結果を\figref{figure:pid-control2}に示す.
  $v_{target}=5$ [km/h]のとき$T^{conv}_{20\%}=10.7$ [sec], $v_{target}=10$ [km/h]のとき$T^{conv}_{10\%}=8.87$ [sec]となっている.
  PID1に比べて大きく速度追従の時間遅れが改善していることがわかる.
}%

\begin{figure}[t]
  \centering
  \includegraphics[width=0.95\columnwidth]{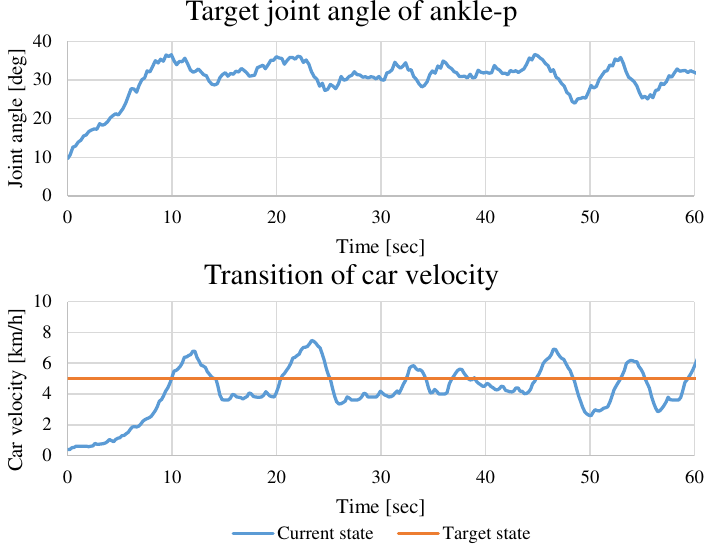}
  \caption{The result of the random controller (\textbf{Random}).}
  \label{figure:random-control}
  \vspace{-3.0ex}
\end{figure}

\begin{figure}[t]
  \centering
  \includegraphics[width=1.0\columnwidth]{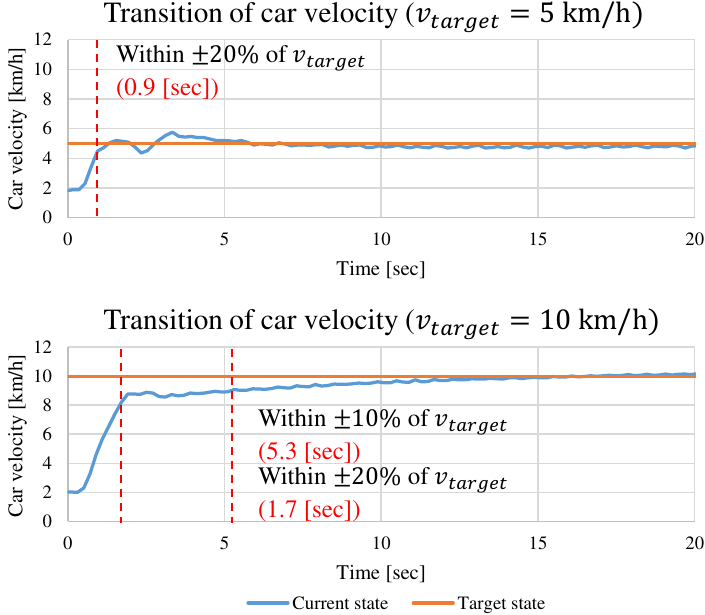}
  \vspace{-3.0ex}
  \caption{The result of the task-specific self-body controller acquisition method (\textbf{Proposed}).}
  \label{figure:mpc-control}
  \vspace{-3.0ex}
\end{figure}

\subsection{Task-specific Self-body Controller Acquisition}
\switchlanguage%
{%
  We show the result of the task-specific self-body controller acquisition method (we call it \textbf{Proposed}).
  At first, we show the control result by the random controller as shown below (we call it \textbf{Random}), in \figref{figure:random-control}.
  \begin{align}
    if\;\;\;\;\;&v(t) \leq v_{target}(t) \notag\\
    &u = u+\textrm{Random}(u^{random}_{min}, u^{random}_{max}) \\
    else\;\;\;& \notag\\
    &u = u-\textrm{Random}(u^{random}_{min}, u^{random}_{max})
  \end{align}
  Where $\textrm{Random}(u^{random}_{min}, u^{random}_{max})$ is the value ranging from $u^{random}_{min}$ to $u^{random}_{max}$.
  In this study, we set $u^{random}_{min}=-1$ [deg] and $u^{random}_{max}=2$ [deg].
  As shown in \figref{figure:random-control}, the ankle joint and car velocity undulate, and we train DDC-Net by using these data.
  We execute this \textbf{Random} for 60 sec, so we can obtain 300 step data, because the control frequency is 5 Hz stated in \secref{subsec:humanoid-self-driving} and $t_{interval}=0.2$ [sec].

  We show the result of \textbf{Proposed} in \figref{figure:mpc-control}.
  $T^{conv}_{20\%}=0.9$ [sec] when $v_{target}=5$ [km/h], and $T^{conv}_{10\%}=5.3$ [sec] when $v_{target}=10$ [km/h].
  Also, $T^{conv}_{20\%}=1.7$ [sec] when $v_{target}=10$ [km/h].
  So, while there are a little error, the car velocity converges to the target task state much better than \textbf{PID1} and \textbf{PID2}.
}%
{%
  本研究で提案したタスク特化型自己身体制御獲得による結果を示す(以降Proposedと呼ぶ).
  まず, 以下のようなランダムコントローラを用いて速度制御を行った結果を\figref{figure:random-control}に示す(以降Randomと呼ぶ).
  \begin{align}
    if\;\;\;\;\;&v(t) \leq v_{target}(t) \notag\\
    &u = u+\textrm{Random}(u^{random}_{min}, u^{random}_{max}) \\
    else\;\;\;& \notag\\
    &u = u-\textrm{Random}(u^{random}_{min}, u^{random}_{max})
  \end{align}
  ここで, $\textrm{Random}(u^{random}_{min}, u^{random}_{max})$は$u^{random}_{min}$から$u^{random}_{max}$の範囲のランダムな値である.
  本研究では$u^{random}_{min}=-1$ [deg], $u^{random}_{max}=2$ [deg]とした.
  \figref{figure:random-control}のように足首関節角度と車速は波打った動きをしており, ここで得たデータを用いてDDC-Netを学習させる.
  この制御はは60秒間行い, \secref{subsec:humanoid-self-driving}で述べた通り5Hzの制御周期のため, $t_{interval}=0.2$ [sec], つまり300ステップ分のデータが得られる.

  上記で得られたデータを学習することで得られたタスク特化型自己身体制御を用いた速度追従実験の結果を\figref{figure:mpc-control}に示す.
  $v_{target}=5$ [km/h]のとき$T^{conv}_{20\%}=0.9$ [sec], $v_{target}=10$ [km/h]のとき$T^{conv}_{10\%}=5.3$ [sec]となっている.
  また, $v_{target}=10$ [km/h]のとき$T^{conv}_{20\%}=1.7$ [sec]となっており, 誤差が多少残ったものの, PID1, PID2と比べて収束はずっと速いことがわかる.
}%

\section{Discussion} \label{sec:discussion}
\switchlanguage%
{%
  First, we consider the experimental results of \secref{sec:experiment}.
  From these experiments, we can see that the convergence to the target velocity is fast in \textbf{Proposed}, compared to in \textbf{PID1} and \textbf{PID2}.
  Although there remains a possibility of improving the convergence time of \textbf{PID1} and \textbf{PID2} by tuning gains, the car velocity vibrates largely when increasing the gain in our experiments.
  \textbf{Proposed} can be constructed only by conducting \textbf{Random} for 60 seconds.
  Although there are some parameters in \textbf{Proposed}, we only had to tune $\alpha$ after setting all parameters.
  When $\alpha$, which is the parameter that limits the changes of control input, is small, the convergence time shortens and the vibration increases, and vice versa.
  Although we set the loss function as \equref{equation:loss} in this study, there are many variations of its setting.
  For example, it is possible to limit joint angles within a certain range or inhibit the acceleration changes of the control input.

  Second, we consider the applicable scope of this study.
  In this study, although we focused on the accelerator pedal control by the musculoskeletal humanoid, there are many other problems to which it can be applied.
  For example, in autonomous driving, this method can be applied to not only the accelerator pedal, but the brake pedal, handle operation, and compensation of body tilt during driving.
  Also, we can change the kind of control input.
  We can use not only the joint angle, but the joint velocity, joint torque, and muscle tension directly.
  Also, although we conducted experiments to track the constant car velocity, we can change the transition of the target car velocity $v_{target}(t)$.
  For example, in the case of applying a brake to smoothly stop at the intended place, we can decide the target transition of the car velocity by minimizing jerk, and conduct a control that tracks it.

  Finally, we consider further developments of \textbf{Proposed}.
  In this study, in accordance with the increase in the dimension of control input, the control input space that needs to be learned increases exponentially, and to handle multi-DOFs is difficult.
  So, we must reduce the control input space by using concepts of muscle synergy \cite{allessandro2013synergy}, AutoEncoder \cite{hinton2006reducing}, etc.
  Also, although we constructed DDC-Net by a simple structure from the viewpoint of stability and cost of learning, we can express this network by recurrent networks such as LSTM \cite{hochreiter1997lstm}.
  Although we start to develop this method in the context of the accelerator pedal control by the musculoskeletal humanoid, there are many other appropriate behaviors and robots for it to be applied.
}%
{%
  まず, \secref{sec:experiment}の実験結果について考察する.
  実験から, PID1, PID2に比べて, Proposedは速度追従の収束が速いことがわかった.
  PID1, PID2は, 更なるゲインチューニングにより収束を良くできる可能性が残るが, 本実験ではこれ以上ゲインを上げると激しく発振してった.
  Proposedは, Randomを60秒間実行するだけで構築することが可能であり, 最適化パラメータは存在するものの, 最初に設定した値から変化させて挙動を確認したパラメータは$\alpha$のみである.
  $\alpha$は制御入力に対する制限の度合いを調整するパラメータであり, 小さくすると収束は速いものの振動が増え, 大きくすると収束は遅くなり振動が抑制される.
  本研究では\equref{equation:loss}のようにlossを設定したが, その設定には様々なバリエーションが考えられ, 関節角度をある範囲に抑えたり, 制御入力の加速度変化を抑制する項を追加することも可能である.

  次に, Proposedの適用範囲に関して考察する.
  本研究では, ヒューマノイドによるアクセルペダル操作に焦点を絞り実験を行ったが, 他にも様々な問題に対して適用可能である.
  例えば, 運転操作ではアクセルだけでなくブレーキやハンドル操作, 運転中の身体の傾き補正等にも応用できると考える.
  また, 制御入力も様々に変更することができると考える.
  関節角度指令ではなく, 関節速度指令, 関節トルク指令, そして, 筋肉の筋張力を直接制御入力とすることも考えられる.
  また, 本研究ではある一定の速度に追従する実験を行ったが, $v_{target}(t)$は時間変化させることができる.
  例えば, 滑らかに所望の場所で止まるようなブレーキをしたい場合, 躍動最小化の観点から時系列の速度指令を決定して, それに追従する制御を実行することができる.

  最後に, Proposedの今後の発展に関して考察する.
  本研究で開発した手法は, 関節軸等の制御入力の数が増加するに従って, 指数関数的に探索する制御入力の空間が増加するため, 多自由度な全身を扱うのは難しい.
  そこで, muscle synergy \cite{allessandro2013synergy}の概念やAutoEncoder \cite{hinton2006reducing}の概念等によって, 制御入力数を削減することが考えられる.
  また, 本研究では学習の安定性やコストの面からDDC-Netを構築しているが, LSTM\cite{hochreiter1997lstm}等のReccurentなネットワークを用いて表現することも可能である.
  本研究は筋骨格ヒューマノイドにおけるアクセルペダル操作の文脈で議論を始めたが, 適した動作・ロボットは無数にあり, 広く適用可能であると考える.
}%

\section{CONCLUSION} \label{sec:conclusion}
\switchlanguage%
{%
  In this study, we proposed the task-specific self-body controller acquisition method, and verified the effectiveness through the accelerator pedal experiments by the musculoskeletal humanoid Musashi.
  It is difficult for the error between the actual robot and its self-body image to completely vanish by merely learning the relationship between joints and muscles, and the task realization is even more difficult.
  Thus, we construct the novel network (DDC-Net) whose input is the current task state and time-series target control input, and whose output is the time-series predicted task state.
  By learning the DDC-Net using the random motion data and backpropagating the error between the network output and intended task state to the target control input, the optimal control input is acquired.
  Using this method, we do not need to tune the PID gain regarding the pedal control, and can realize the intended car velocity stably and dynamically.
  In future works, we would like to analyze the stability of this neural network based controller for the actual autonomous driving.
  Also, we would like to realize long series consecutive tasks by the musculoskeletal humanoid.
}%
{%
  本研究では, 筋骨格ヒューマノイドによる車のペダル操作を例に, タスク特化型の自己身体制御獲得手法について述べた.
  関節と筋の非線形な関係を学習するだけでは自己身体像と実機の誤差を完全に無くすことが難しいため, 実際のタスク実現はより困難となる.
  そこで, 現タスク状態から制御入力の時系列情報を入力し, 次タスク状態の時系列を出力するようなネットワークを構築する.
  これをランダムな動作データから学習し, 所望のタスク状態との誤差から入力に対する誤差逆伝播を用いることで最適な時系列制御入力を獲得可能な手法を開発した.
  これにより, ペダル操作においてPID制御等におけるゲインチューニングや最適制御におけるモデリングが不要となり, 動的に所望の車速を安定して実現できることを確認した.
  今後は, モデリング困難な筋骨格ヒューマノイドにより, 大きな一連のタスクを動的に実現する方法について考えたい.
}%

{
  \bibliographystyle{IEEEtran}
  \bibliography{main}
}

\end{document}